\documentclass[10pt,twocolumn,letterpaper]{article}
\usepackage[margin=0.75in,columnsep=0.25in]{geometry}
\usepackage{times,graphicx,amsmath,amssymb,booktabs,xcolor}
\usepackage[numbers,sort&compress]{natbib}
\usepackage[font=small,labelformat=simple,labelsep=period]{caption}
\usepackage{subcaption}
\date{}
\usepackage[T1]{fontenc}
\usepackage{microtype}
\usepackage{xurl,colortbl,tabularx,adjustbox}
\usepackage{algorithm,algpseudocode,placeins}
\definecolor{cvprblue}{rgb}{0.21,0.49,0.74}
\usepackage[pagebackref,breaklinks,colorlinks,allcolors=cvprblue]{hyperref}
\title{Cognitive-Horizon Action Chunking:\\Memory-Guided Adaptive Reasoning for Embodied Navigation}
\author{\begin{tabular}{c}
Weile Gong$^{1}$, Zijian Lu$^{1,\dagger}$, Jiahao Zheng$^{1,\dagger}$, Bo Xu$^{1}$, Yiping Zuo$^{1}$\\
Jiajia Guo$^{2}$, Xin He$^{1}$, Weijun Wang$^{3}$, Weibei Fan$^{1}$\\[6pt]
\small $^{1}$Nanjing University of Posts and Telecommunications\\
\small $^{2}$National Mobile Communications Research Laboratory, Southeast University\\
\small $^{3}$Institute for AI Industry Research, Tsinghua University\\[6pt]
\footnotesize\texttt{phare111@163.com, 18818732360@163.com, jiahaozheng2005@163.com}\\
\footnotesize\texttt{1226044921@njupt.edu.cn, zuoyiping@njupt.edu.cn, jiajiaguo@seu.edu.cn}\\
\footnotesize\texttt{xhe@njupt.edu.cn, wangweijun@air.tsinghua.edu.cn, wbfan@njupt.edu.cn}\\[4pt]
\small $^{\dagger}$Co-second authors.
\end{tabular}}
\hypersetup{pdftitle={Cognitive-Horizon Action Chunking: Memory-Guided Adaptive Reasoning for Embodied Navigation},pdfauthor={Weile Gong, Zijian Lu, Jiahao Zheng, Bo Xu, Yiping Zuo, Jiajia Guo, Xin He, Weijun Wang, Weibei Fan}}
\begin{document}
\maketitle
\raggedbottom
\begin{abstract}
Multimodal foundation models are advancing embodied navigation, but frequent inference incurs substantial computational cost. We present Cognitive-Horizon Action Chunking (CHAC), a training-free framework for memory-guided adaptive reasoning with a frozen navigation policy. Retained visual context, completion cues, and controller feedback determine the cognitive horizon, or how many cached waypoints the robot executes before refreshing its prediction. At the policy's stop request, local visitation history guides a bounded correction with at most one extra movement and no additional model calls. CHAC reduces vision-language model calls by 33.4\% on HM3D-OVON and 64.7\% on R2R-CE, with success-rate decreases of 0.63 and 10.82 percentage points, respectively. Success rate and success weighted by path length improve per model call on both benchmarks. Across four physical navigation tasks with 40 runs per method, CHAC achieves 18 successes versus 14 for native execution, reducing model calls by 54.9\% and GPU inference time by 49.7\%.
\par\noindent Code: \url{https://github.com/phare111/chac}.
\end{abstract}
\section{Introduction}
\begin{figure*}[!t]
\centering
\includegraphics[width=\textwidth,trim=0 37bp 0 38bp,clip]{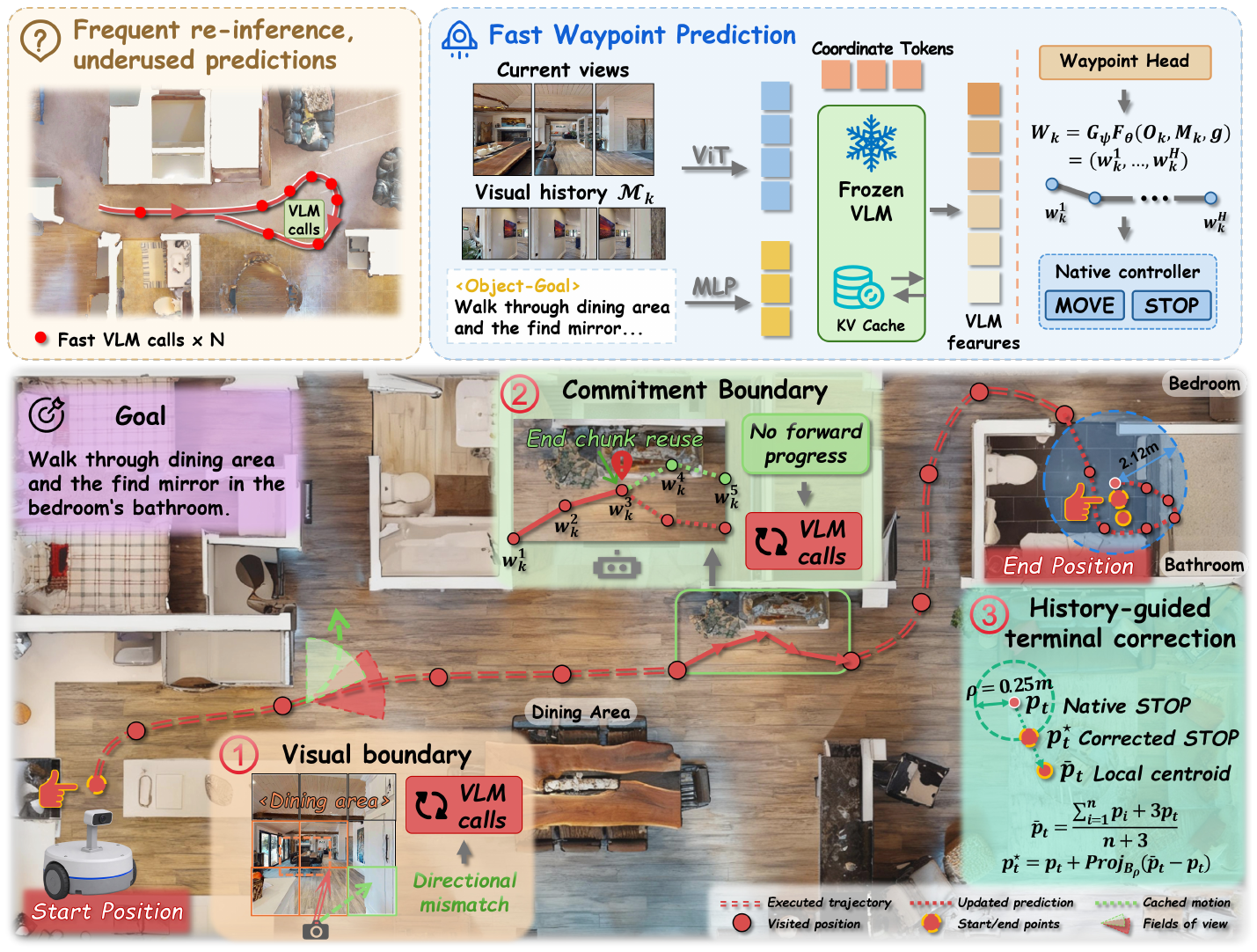}
\caption{\textbf{Overview of CHAC.} A frozen navigation policy predicts ordered waypoints from current views, visual history, and the task instruction. During execution, retained visual context and commitment checks guide when to refresh the cached prediction. At native STOP, nearby visited positions guide a bounded terminal correction without another model call. The lower scene shows how executed motion becomes visitation history. Red dots mark visited positions, and the blue disk denotes the local-history neighborhood.}
\label{fig:teaser}
\end{figure*}
Recent advances in multimodal foundation models are expanding embodied intelligence. Robot-control evaluations of general-purpose multimodal models~\cite{anthropic2026fable52} and advances in manipulation with Gemini Robotics~2.5~\cite{parada2026geminirobotics25} illustrate the promise of connecting visual understanding and language instructions with embodied action. For navigation, vision-language models (VLMs) interpret scenes, ground goals and instructions, and generate movement through unfamiliar environments~\cite{zhang2024navidvideobasedvlmplans,xue2026omninav,hong2026simplevlnsimplefrontiervisionlanguage,zhu2026textscnavidavisionlanguagenavigationinverse}. Sustained navigation requires coordinating this inference with physical execution.

Repeatedly invoking large vision-language policies incurs computational cost over many physical actions. Action chunking amortizes this cost by predicting several future actions per invocation~\cite{zhao2023learningfinegrainedbimanualmanipulation,chi2024diffusionpolicyvisuomotorpolicy}. Its benefit depends on execution: frequent refreshes limit reuse, while extended execution can carry a prediction beyond its supporting observations and motion conditions. Efficient navigation requires deciding how long to act on a prediction and when new evidence warrants another policy invocation.

Existing approaches adjust inference timing or execution horizons using consistency or uncertainty signals~\cite{so2025improving,huang2026chunktrustadaptingexecutionhorizons,liang2026adaptiveactionchunkinginferencetime,pmlr-v306-ding26r,akustia2026aspen}. Navigation must also account for movement-induced changes: a turn reveals a new action direction, a completion cue warrants fresh inference, or the controller cannot realize the requested motion. Refreshing changes subsequent actions, observations, and execution before termination. Its computational effect therefore extends beyond the immediate cost of a new call, motivating decisions based on current evidence and retained context.

Our key insight is that retained visual context and visitation history can guide both inference timing and terminal execution. Visual references and observation coverage preserve the context available when each prediction was generated. Completion cues and motion feedback complement this context when deciding whether to reuse or refresh a prediction. Executed actions also leave a position history that remains available after prediction replacement. Our offline trajectory analysis reveals episodes in which the robot enters a suitable stopping region but later stops outside it (Section~\ref{sec:terminal-correction-results}). Nearby visited positions can therefore provide a geometric reference for a bounded terminal correction.

We introduce \textbf{Cognitive-Horizon Action Chunking (CHAC)}, a training-free framework for \emph{memory-guided adaptive reasoning} with a frozen navigation policy (Fig.~\ref{fig:teaser}). CHAC determines when to reuse a cached prediction and when to invoke the policy again. Its \emph{cognitive horizon} counts the ordered waypoints executed before refresh. A visual boundary compares appearance along newly observed action directions with retained visual context. A commitment boundary checks completion cues and controller-supported progress. Together with inherited motion feedback, they determine inference timing. At native STOP, local visitation history guides a bounded correction, adding at most one movement and no model calls.

Our contributions are threefold:
\begin{itemize}
\item We introduce CHAC, a training-free framework for memory-guided adaptive reasoning combining an online cognitive horizon for policy invocation with history-guided terminal correction.
\item We identify how refresh decisions interact with subsequent execution to determine inference demand, including a reversal of visual refresh's computational effect under commitment checks.
\item We evaluate navigation quality and inference cost across two benchmarks and physical tasks, with matched terminal interventions testing visitation history's directional information.
\end{itemize}

\section{Related Work}

\subsection{Vision-Language-Action Models for Navigation}

Vision-and-language navigation (VLN) grounds language in sequential visual decisions~\cite{anderson2018visionandlanguagenavigationinterpretingvisuallygrounded}, while vision-language-action (VLA) policies map multimodal observations and instructions to executable actions~\cite{pmlr-v229-zitkovich23a,pmlr-v270-kim25c}. NaVid uses observation history to predict actions~\cite{zhang2024navidvideobasedvlmplans}. OmniNav, SimpleVLN, and NaVIDA predict waypoints or action chunks~\cite{xue2026omninav,hong2026simplevlnsimplefrontiervisionlanguage,zhu2026textscnavidavisionlanguagenavigationinverse}. ETPNav organizes waypoints in an evolving topological map~\cite{10495141}. These models supply multi-step predictions whose execution and refresh affect deployment cost.

\subsection{Action Chunking and Adaptive Reasoning}

ACT combines multi-step prediction with temporal ensembling~\cite{zhao2023learningfinegrainedbimanualmanipulation}, and Diffusion Policy generates actions through conditional denoising~\cite{chi2024diffusionpolicyvisuomotorpolicy}. BID selects chunks using bidirectional consistency~\cite{liu2025bidirectional}. SGAC combines self-guidance with agreement between predictions~\cite{so2025improving}. Adaptive horizons use action entropy, attention, cross-horizon agreement~\cite{liang2026adaptiveactionchunkinginferencetime,wang2026vlaknowslimitsadaptive,pmlr-v306-jing26f}, denoising variance in DVAC~\cite{feng2026denoisingtellsreplandenoisingvariance}, trajectory geometry in GeoAAC~\cite{chen2026geoaacgeometrybasedadaptiveaction}, or action-expert stability in ChunkTrust~\cite{huang2026chunktrustadaptingexecutionhorizons}. A$^3$ verifies candidate prefixes~\cite{chen2026dynamicexecutioncommitmentvisionlanguageaction}, while DEHP learns a horizon with a frozen policy~\cite{zhao2026dynamicexecutionhorizonprediction}. AdaNav learns uncertainty-based reasoning triggers~\cite{pmlr-v306-ding26r}. ASPEN learns command persistence to schedule reuse or fresh inference in a frozen navigation VLA~\cite{akustia2026aspen}. CHAC instead regulates ordered waypoint reuse with retained visual context and execution checks, without training a scheduler.

\subsection{Memory and Feedback-Guided Execution}

MEM combines short-term video and long-term language memory for robot control~\cite{torne2026mem}. In navigation, EvoNav summarizes past trajectories~\cite{Dai_2026_CVPR}, and MSGNav selects a final viewpoint using visibility~\cite{Huang_2026_CVPR}. Feedback also guides inference and correction: RTC and LiveVLN overlap inference with motion~\cite{NEURIPS2025_300ccb21,11694288}, Slow4fast-VLN coordinates slow reasoning with control~\cite{li2026openenvironmentsinstructionsgeneral}, and A2C2 corrects cached actions~\cite{sendai2025leaveobservationbehindrealtime}. Scene-change triggers adapt inference timing~\cite{wu2026reactneedtoeventtriggered}. For behavioral correction, CorrectNav learns from error trajectories~\cite{Yu_Long_Yang_Zeng_Fan_Zhang_Dong_2026}, and SC$^2$-WM uses world-model feedback~\cite{pmlr-v306-yao26u}. CHAC uses existing visual context to schedule policy calls and local visitation geometry for terminal correction, with frozen model weights.

\section{Preliminaries}
\label{sec:prelim}
\paragraph{OmniNav-Fast.}
OmniNav~\cite{xue2026omninav} combines fast waypoint prediction with slower planning. We use its standalone fast system, which generates continuous waypoints through a flow-matching head conditioned on vision-language features. Given a goal $g$, three current red-green-blue (RGB) views $\mathcal O_k$, and native visual history $\mathcal M_k$, inference $k$ produces an ordered waypoint sequence:
\begin{equation}
\mathcal W_k=G_\psi\bigl(F_\theta(\mathcal O_k,\mathcal M_k,g)\bigr)
             =(w_k^1,\ldots,w_k^H).
\label{eq:policy}
\end{equation}
Here, $H=5$ is the number of predicted waypoints. The vision-language backbone $F_\theta$ and flow-matching head $G_\psi$ have frozen parameters $\theta$ and $\psi$, respectively. Each waypoint $w_k^i=(x_k^i,y_k^i,\sin\theta_k^i,\cos\theta_k^i,c_k^i)$ encodes planar position $(x_k^i,y_k^i)$, heading $\theta_k^i$, and arrival score $c_k^i$, with $i=1,\ldots,H$. The native interface requests the stop action (STOP) when $c_k^i>0.5$ for every waypoint. Otherwise, it decodes the first waypoint into a motion command.

\paragraph{Action reuse.}
Execution uses an ordered prefix of the predicted chunk, with length $h_k\leq H$ defining its cognitive horizon. Native execution refreshes the prediction after executing the first waypoint. CHAC determines this prefix online while preserving waypoint order and native decoding. OmniNav reports five Euler integration steps~\cite{xue2026omninav}. Our frozen evaluation configuration uses ten flow-head updates per chunk, with VLM calls as the principal measure of inference demand.

\section{Cognitive-Horizon Action Chunking}
\label{sec:method}
CHAC implements memory-guided adaptive reasoning through the online scheduling of frozen-policy invocations. Retained visual references and observation coverage inform prediction reuse. Accumulated position records provide a geometric reference for terminal correction. New visual and motion evidence determines whether the remaining waypoints are executed or refreshed. Refreshing replaces the prediction while retaining navigation history. At the native STOP request, nearby visited positions supply the geometric reference for a bounded correction.

\subsection{Memory-Guided Action Reuse}
CHAC retains a prediction while its execution checks permit continued motion. The \emph{cognitive horizon} $h_k$ describes this lifetime as observations and motion feedback evolve. After waypoint $j$, a refresh event $B_{k,j}$ ends reuse before the next cached movement. The available information $\mathcal F_{k,j}$ includes the prediction, observations, execution state, motor feedback, and native read-only controller previews. The first such event determines the executed prefix:
\begin{equation}
 h_k=\min\!\left\{H,\inf\{j\geq1:B_{k,j}\}\right\}.
\label{eq:horizon}
\end{equation}
Boundary checks are restricted to $j<H$, with $\inf\varnothing=+\infty$: absent a refresh event, the full chunk is reused. Each event uses information available at the check, making $h_k$ a bounded stopping time with $\{h_k\leq j\}\in\mathcal F_{k,j}$ (Appendix~\ref{app:math}). The episode budget imposes an additional execution limit.

\paragraph{Commitment boundary.}
The commitment boundary checks predicted completion and whether the controller supports the next motion. After waypoint $j$, an arrival score $c_k^j>0.5$ prompts a refresh before a subsequent MOVE. This cue triggers a new prediction. The native STOP rule continues to govern episode termination. The controller also previews the next requested displacement $\mathbf u$ and its filtered, snapped outcome $\widehat{\mathbf v}$. For a nonzero MOVE request, supported progress along the requested direction is
\begin{equation}
\gamma_{k,j}=\frac{\mathbf u^\top\widehat{\mathbf v}}{\|\mathbf u\|_2}.
\label{eq:commitment}
\end{equation}
A nonpositive $\gamma_{k,j}$ indicates no forward progress along the request and triggers a refresh. Either a completion cue or nonpositive previewed progress activates the commitment event $C_{k,j}$. Both displacements lie in the world horizontal plane. The read-only preview uses native controller geometry without executing environment actions, preserving simulator pose and random state.

\paragraph{Visual boundary.}
Movement can expose a new view along the next predicted action. CHAC checks whether appearance in that direction differs from the visual context supporting the prediction. At inference $k$, it stores the angular coverage $\Omega_k$ of the current views and native visual history, together with three $32\times32$ grayscale references. Current coverage $\Omega_t$ follows the camera configuration and command odometry.

For a visible cached MOVE direction $\phi_t$, CHAC selects the current camera closest in angle. Its image and the three references are resampled to $32$ rows and $30$ columns and split into three vertical regions each. Let $R_{t,q}$ denote a current region, $A_{k,s}$ one of the nine reference regions, and $q^\star$ the region containing $\phi_t$. With $P=320$ pixels per region, we first find each region's closest reference and then contrast the action region with the other current regions:
\begin{equation}
\begin{aligned}
 d_{t,q}&=\min_{1\leq s\leq9}\frac{\|R_{t,q}-A_{k,s}\|_1}{P},\\
 \delta_{k,j}&=d_{t,q^\star}-\max_{q\ne q^\star}d_{t,q}.
\end{aligned}
\label{eq:appearance}
\end{equation}
A positive $\delta_{k,j}$ means that the action region differs from its closest reference more than either other region differs from its own match. The visual event $V_{k,j}$ triggers refresh only when this contrast is positive and the action direction is newly observed: $\phi_t\in\Omega_t\setminus\Omega_k$ and $\delta_{k,j}>0$. The reference remains fixed within the chunk.

\paragraph{Coordinated execution.}
The two boundaries check the same prediction before further reuse. Before each cached MOVE, CHAC checks directional visual contrast, predicted arrival, and controller-supported progress in sequence, refreshing at the first trigger. After execution, inherited motor feedback triggers $M_{k,j}$ upon collision, rejection, or nonpositive realized progress $\mathbf u^\top\mathbf v\leq0$ for a nonzero request, where $\mathbf v$ is the realized displacement. These events invalidate the cache and jointly determine $B_{k,j}=C_{k,j}\cup V_{k,j}\cup M_{k,j}$ and the horizon in Eq.~\eqref{eq:horizon}.

For fixed execution evidence, an additional boundary can only preserve or shorten the current horizon. Over a closed-loop episode, however, refresh changes subsequent actions, observations, and stopping decisions. Total calls depend on both calls per action and the number of actions executed. A shorter horizon therefore need not increase total calls. Section~\ref{sec:boundary-coordination} examines this distinction through paired runs, including savings associated with earlier failed stops.

\subsection{History-Guided Geometric Projection}
\label{sec:terminal-projection}
The same execution process accumulates positions that remain available after individual predictions are refreshed. At a native STOP request, CHAC uses nearby positions as a local geometric reference for terminal correction. It retains visual-history positions within distance $Hb$ of the current position $\mathbf p_t$. The multiset $\mathcal L_t$ preserves their native multiplicity and includes three copies of $\mathbf p_t$, one per current view. Native first-waypoint translation bounds $\mathbf m_1,\mathbf M_1$ give the history radius $Hb$ and correction bound $\rho$:
\begin{equation}
\begin{aligned}
 b&=s_0\|\max(|\mathbf m_1|,|\mathbf M_1|)\|_2,\\
 \rho&=\min(b,s_{\mathrm{motor}}).
\end{aligned}
\label{eq:scale}
\end{equation}
The maximum is elementwise. The native scales $s_0=0.3$ and $s_{\mathrm{motor}}=0.25\,\mathrm{m}$ give $b\approx0.424\,\mathrm{m}$ and $\rho=0.25\,\mathrm{m}$.

Let $\mathbf y_i=[R_t^\top(\mathbf p_i-\mathbf p_t)]_{xz}$ be the horizontal offset of each retained position, with $R_t$ mapping local to world coordinates. These records contain no goal labels. The correction seeks agreement with their local geometry under a motion bound:
\begin{equation}
 \boldsymbol\Delta_t^\star
 =\underset{\|\boldsymbol\Delta\|_2\leq\rho}{\arg\min}
   \frac1{|\mathcal L_t|}\sum_i\|\boldsymbol\Delta-\mathbf y_i\|_2^2.
\label{eq:terminal}
\end{equation}
\begin{samepage}
Up to a constant, the objective is $\|\boldsymbol\Delta-\overline{\mathbf y}_t\|_2^2$, where $\overline{\mathbf y}_t=|\mathcal L_t|^{-1}\sum_i\mathbf y_i$. Its unique minimizer projects this mean onto the disk $\mathbb B_\rho=\{\mathbf d:\|\mathbf d\|_2\leq\rho\}$ and stays in the convex hull of the retained offsets (Appendix~\ref{app:math}):
\begin{equation}
\begin{aligned}
 \boldsymbol\Delta_t^\star
 &=\operatorname{Proj}_{\mathbb B_\rho}(\overline{\mathbf y}_t),\\
 \boldsymbol\Delta_t^\star
 &\in\operatorname{conv}\{\mathbf0,\mathbf y_1,\ldots,\mathbf y_n\}
       \cap\mathbb B_\rho.
\end{aligned}
\label{eq:projection}
\end{equation}
\end{samepage}
For $n>0$ historical offsets, the three current-view zeros shrink the mean to $\overline{\mathbf y}_t=\frac{n}{n+3}\overline{\mathbf y}_{\mathrm{hist}}$. For $n=0$, the candidate is zero. The projection bounds the requested correction. Section~\ref{sec:terminal-correction-results} tests its navigation benefit through matched interventions.

CHAC converts the candidate into a native MOVE with distance $\|\boldsymbol\Delta_t^\star\|_2$ and heading $\operatorname{atan2}(-\Delta_{t,x}^\star,-\Delta_{t,z}^\star)$. With at least two action slots remaining, a nonzero candidate that passes the positive preview criterion in Eq.~\eqref{eq:commitment} precedes the original STOP. Otherwise, STOP executes directly. The correction adds at most one MOVE and zero model calls.

\paragraph{Unified execution.}
Algorithm~\ref{alg:chac} integrates reuse, refresh, and terminal correction in one execution loop. A refresh replaces unexecuted waypoints and renews the visual reference while retaining navigation history. Accepted movements update that history and receive motor feedback. A native STOP request invokes the bounded correction, subject to the conditions above, before stopping. Native decoding uses command odometry. Available motion-state estimates provide feedback and populate the spatial history.

\begin{algorithm}[!t]
\caption{Cognitive-Horizon Action Chunking}
\label{alg:chac}
\small
\begin{algorithmic}[1]
\Require Frozen policy $\pi$ from Eq.~\eqref{eq:policy}, goal $g$, native controller
\State Initialize an empty cache, navigation histories, and $k\gets0$
\While{episode active and budget remains}
  \If{cache invalid or exhausted}
    \State $k\gets k+1$
    \State $\mathcal W_k\gets\pi(\mathcal O_k,\mathcal M_k,g)$, $j\gets1$, anchor visual reference
  \EndIf
  \If{native STOP requested}
    \State Form $\mathcal L_t$ and $\overline{\mathbf y}_t$ as in Section~\ref{sec:terminal-projection}
    \State $\boldsymbol\Delta_t^\star\gets\operatorname{Proj}_{\mathbb B_\rho}(\overline{\mathbf y}_t)$
    \State Apply correction if nonzero, remaining action budget $\ge2$, and preview supports positive progress (Eq.~\eqref{eq:commitment})
    \State STOP and \textbf{break}
  \EndIf
  \State $a\gets\operatorname{Decode}(w_k^j)$
  \If{$j>1$ and $C_{k,j-1}\cup V_{k,j-1}$ occurs}
    \State Invalidate cache and \textbf{continue}
  \EndIf
  \State Execute $a$, update histories, and set $j\gets j+1$
  \State Invalidate cache if motor feedback triggers $M_{k,j-1}$
\EndWhile
\end{algorithmic}
\end{algorithm}

\newcommand{\hmrows}{%
OmniNav-Fast~\cite{xue2026omninav} (ICLR 2026) & \textbf{32.17} & \textbf{21.92} & 130,444 & 1,304,440 & 130,444 & \underline{1.000} & \underline{1.000} \\
AAC~\cite{liang2026adaptiveactionchunkinginferencetime} (CVPR 2026) & 14.20 & 5.28 & 84,869 & 16,973,800 & 276,726 & 0.679 & 0.376 \\
SGAC~\cite{so2025improving} (NeurIPS 2025) & \textit{26.23} & \textit{16.18} & 330,314 & $\geq$3,303,140 & 330,314 & 0.322 & 0.300 \\
ChunkTrust-AHS~\cite{huang2026chunktrustadaptingexecutionhorizons} (arXiv 2026) & 17.13 & 7.72 & 98,596 & 985,960 & 207,696 & 0.705 & 0.476 \\
DVAC~\cite{feng2026denoisingtellsreplandenoisingvariance} (arXiv 2026) & 7.13 & 1.71 & 91,006 & 910,060 & 350,266 & 0.318 & 0.112 \\
GeoAAC~\cite{chen2026geoaacgeometrybasedadaptiveaction} (arXiv 2026) & 13.77 & 4.35 & 99,317 & 993,170 & 297,977 & 0.562 & 0.263 \\
\rowcolor{gray!12}
\textbf{CHAC (Ours)} & \underline{31.53} & \underline{16.82} & 86,846 & 868,460 & 280,483 & \textbf{1.472} & \textbf{1.198} \\
}
\newcommand{\rtrrows}{%
OmniNav-Fast (ICLR 2026) & \textbf{57.04} & \textbf{49.30} & 240,526 & 2,405,260 & 240,526 & 1.000 & 1.000 \\
AAC (CVPR 2026) & 36.70 & 23.02 & 35,236 & 7,047,200 & 119,000 & \textbf{4.392} & \underline{3.187} \\
SGAC (NeurIPS 2025) & \underline{55.46} & \underline{46.17} & 255,276 & 4,720,110 & 255,276 & 0.916 & 0.882 \\
ChunkTrust-AHS (arXiv 2026) & 46.11 & \textit{34.68} & 51,049 & 510,490 & 110,494 & \underline{3.809} & \textbf{3.314} \\
DVAC (arXiv 2026) & 17.29 & 9.52 & 36,128 & 361,280 & 132,847 & 2.018 & 1.286 \\
GeoAAC (arXiv 2026) & 34.58 & 22.05 & 40,222 & 402,220 & 119,814 & 3.626 & 2.674 \\
\rowcolor{gray!12}
\textbf{CHAC (Ours)} & \textit{46.22} & 34.17 & 84,989 & 849,890 & 347,054 & 2.293 & 1.961 \\
}
\newcommand{\ablationrows}{%
OmniNav-Fast (Native) & 30.60 & 20.47 & 45,286 & 452,860 & 45,286 & 1.000 & 1.000 \\
Fixed-3 & 25.20 & 15.10 & 39,683 & 396,830 & 96,883 & 0.940 & 0.842 \\
Fixed-5 & 21.40 & 11.60 & 33,814 & 338,140 & 116,014 & 0.937 & 0.766 \\
\specialrule{0.3pt}{0.5pt}{0.5pt}
Fixed-5 + Visual & 21.00 & 11.46 & 35,111 & 351,110 & 118,311 & 0.885 & 0.728 \\
Fixed-5 + Commitment & 23.10 & 13.13 & 30,863 & 308,630 & 96,878 & 1.108 & 0.949 \\
Fixed-5 + Terminal & 29.40 & 14.77 & 33,814 & 338,140 & 116,943 & 1.287 & 0.975 \\
\specialrule{0.3pt}{0.5pt}{0.5pt}
w/o Visual Boundary & 29.40 & 15.75 & 30,863 & 308,630 & 97,833 & 1.410 & 1.139 \\
w/o Commitment Boundary & 29.10 & 14.74 & 35,111 & 351,110 & 119,236 & 1.227 & 0.937 \\
w/o Terminal Correction & 23.00 & 13.11 & 29,811 & 298,110 & 94,369 & 1.142 & 0.982 \\
\specialrule{0.3pt}{0.5pt}{0.5pt}
Full CHAC & 29.60 & 15.93 & 29,811 & 298,110 & 95,328 & 1.469 & 1.193 \\
}

\raggedbottom
\section{Experiments}
\label{sec:experiments}
We evaluate how prior inference remains useful during cached execution and as visitation history. Benchmark comparisons establish the trade-off between navigation quality and inference cost. Ablations and paired analyses examine boundary coordination and terminal correction, followed by physical trials of prospective reuse.

\begin{table*}[!t]
\centering
\caption{\textbf{Inference demand and navigation quality.} Costs are totals over each benchmark. Gray rows identify CHAC. Bold, underlined, and italic entries identify the top three SR/SPL values per benchmark. For $E_{\rm SR}$ and $E_{\rm SPL}$, bold and underline identify the top two. SPL and $E_{\rm SPL}$ use the common valid subset. $E_{\rm SR}$ uses all episodes.}
\label{tab:main}
\small
\setlength{\tabcolsep}{3pt}
\renewcommand{\arraystretch}{1.12}
\begin{tabularx}{\textwidth}{Xrrrrrrr}
\toprule
Method & SR (\%) $\uparrow$ & SPL (\%) $\uparrow$ & VLM Calls $\downarrow$ & FlowEq $\downarrow$ & Env. Actions & $E_{\rm SR}$ $\uparrow$ & $E_{\rm SPL}$ $\uparrow$\\
\midrule
\multicolumn{8}{@{}l}{\textit{HM3D-OVON Val-Unseen}~\cite{yokoyama2024hm3dovon}}\\[2pt]
\hmrows
\midrule
\multicolumn{8}{@{}l}{\textit{R2R-CE Val-Unseen}~\cite{krantz2020beyondnavgraph}}\\[2pt]
\rtrrows
\bottomrule
\end{tabularx}
\end{table*}

\subsection{Experimental Setup}
\paragraph{Benchmarks and policy.}
We evaluate the frozen OmniNav-Fast policy~\cite{xue2026omninav} on HM3D-OVON Val-Unseen~\cite{yokoyama2024hm3dovon} with 3,000 object-goal episodes and R2R-CE Val-Unseen~\cite{krantz2020beyondnavgraph} with 1,839 instruction-following episodes. Success requires an actual STOP with final distance below 0.25\,m and 3\,m, respectively. Both evaluations allow 500 environment actions per episode. Component analyses use a fixed 1,000-episode HM3D development subset, with terminal-direction interventions sharing neural execution prefixes. We fixed CHAC's configuration before evaluating it on R2R-CE. The frozen policy handles goal search and route following. CHAC determines how long to reuse its predictions.

\paragraph{Comparisons.}
We compare native first-waypoint execution of OmniNav-Fast~\cite{xue2026omninav} with adaptations of AAC~\cite{liang2026adaptiveactionchunkinginferencetime}, ChunkTrust-AHS~\cite{huang2026chunktrustadaptingexecutionhorizons}, DVAC~\cite{feng2026denoisingtellsreplandenoisingvariance}, GeoAAC~\cite{chen2026geoaacgeometrybasedadaptiveaction}, and SGAC~\cite{so2025improving} (Table~\ref{tab:main}). The baseline implementations retain their respective observation-history and decoding interfaces. Component studies compare fixed horizons, individual mechanisms, and their combinations on matched episode identities. All horizon variants retain native motor feedback.

\paragraph{Metrics.}
We report success rate (SR), success weighted by path length (SPL), total VLM calls, flow-update equivalents (FlowEq), and environment actions. FlowEq includes recorded additional sampling, with a recorded lower bound for HM3D SGAC. To summarize quality per VLM invocation, we normalize to native execution:
\begin{equation}
 E_Q=
 \frac{\sum_{i\in J_Q}q_i\,/\,\sum_{i\in J_Q}c_i}
      {\sum_{i\in J_Q}q_{0,i}\,/\,\sum_{i\in J_Q}c_{0,i}},
 \quad Q\in\{\mathrm{SR},\mathrm{SPL}\}.
\label{eq:efficiency}
\end{equation}
Here, $q_i$ is success or SPL, $c_i$ counts VLM calls, and subscript $0$ denotes native execution on the same cohort. $J_{\rm SR}$ includes all episodes. $J_{\rm SPL}$ is the common finite-SPL subset: 2,964 HM3D episodes, all 1,839 R2R episodes, and 991 ablation episodes. Table costs cover the complete cohorts. $E_{\rm SPL}$ uses calls from its corresponding valid subset.

\begin{table*}[!t]
\centering
\caption{\textbf{Component contributions on 1,000 matched HM3D episodes.} Visual, Commitment, and Terminal denote the two refresh boundaries and terminal correction. Single-component rows extend Fixed-5. Removal rows start from Full CHAC. SPL uses 991 common valid episodes.}
\label{tab:ablation}
\small
\setlength{\tabcolsep}{4pt}
\renewcommand{\arraystretch}{1.12}
\begin{tabularx}{\textwidth}{Xrrrrrrr}
\toprule
Configuration & SR (\%) $\uparrow$ & SPL (\%) $\uparrow$ & VLM Calls $\downarrow$ & FlowEq $\downarrow$ & Env. Actions & $E_{\rm SR}$ $\uparrow$ & $E_{\rm SPL}$ $\uparrow$\\
\midrule
\ablationrows
\bottomrule
\end{tabularx}
\end{table*}

\begin{figure*}[!t]
\centering
\includegraphics[width=\textwidth]{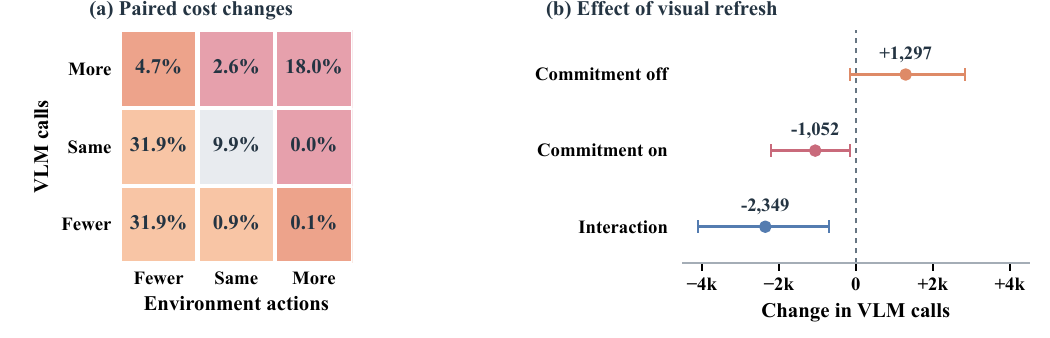}
\caption{\textbf{Global cost changes and boundary interaction.} (a) Full CHAC versus w/o Commitment on all 1,000 pairs. Both retain Visual and Terminal. Peach indicates lower cost, rose indicates higher cost, coral indicates mixed changes, and gray indicates unchanged costs. (b) Adding Visual with Commitment off or on, and the difference between these effects. All four configurations disable Terminal. Error bars are paired-bootstrap 95\% intervals.}
\label{fig:boundary-results}
\end{figure*}

\begin{figure*}[!t]
\centering
\includegraphics[width=\textwidth]{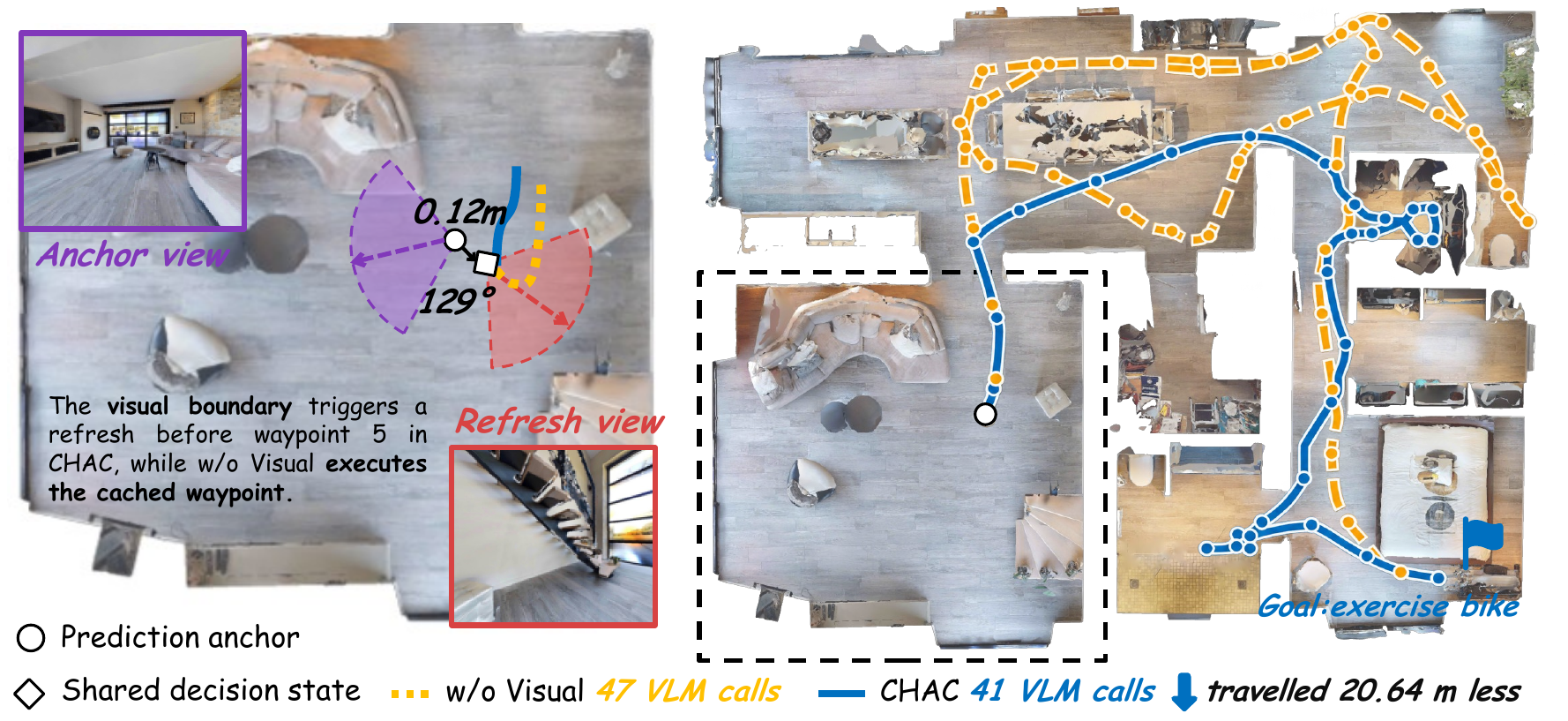}
\caption{\textbf{Small displacement, changed visual evidence (episode 1644).} Blue solid: CHAC; orange dashed: w/o Visual. Both retain Commitment and Terminal. Left: the anchor and refresh states are only 0.12\,m apart but differ in heading by $129^\circ$; purple and red mark the corresponding views. CHAC refreshes before waypoint 5, while w/o Visual executes the cached waypoint. Right: full trajectories toward the exercise bike, with dots marking VLM calls. CHAC uses 41 calls versus 47 and travels 20.64\,m less.}
\label{fig:reorientation}
\end{figure*}

\subsection{Main Results}
On HM3D-OVON, CHAC preserves near-native task completion with substantially fewer model calls (Table~\ref{tab:main}). It achieves 946 successes versus 965 for native execution, a 0.63-percentage-point decrease in SR, while reducing VLM calls by \textbf{33.4\%}, from 130,444 to 86,846. This saving accompanies lower path efficiency: SPL falls from 21.92\% to 16.82\%, and environment actions rise from 130,444 to 280,483. CHAC achieves the highest SR and SPL among the evaluated action-chunking alternatives. At similar VLM call counts, CHAC and AAC achieve 31.53\% and 14.20\% SR, respectively.

The fixed-configuration R2R-CE evaluation shows a larger navigation cost. CHAC reduces calls by \textbf{64.7\%}, from 240,526 to 84,989, while SR falls from 57.04\% to 46.22\% and SPL from 49.30\% to 34.17\%. ChunkTrust-AHS achieves a similar SR of 46.11\% with 51,049 calls and 34.68\% SPL. Thus, inference savings persist across tasks, while the extent of navigation-quality preservation and the competitive trade-off are task dependent.

Relative to native execution, CHAC achieves $E_{\rm SR}=1.472$ and $E_{\rm SPL}=1.198$ on HM3D, and 2.293 and 1.961 on R2R. These ratios report navigation quality per VLM call at each method's evaluated configuration.

\subsection{Ablation Study}
Table~\ref{tab:ablation} reveals complementary contributions during execution and at stopping. Moving from native execution to Fixed-5 reduces calls from 45,286 to 33,814, while SR falls from 30.60\% to 21.40\%. Adding terminal correction restores SR to 29.40\% and raises SPL from 11.60\% to 14.77\% without additional model calls. The executed history therefore supports successful stopping beyond the use of predictions as cached actions.

With terminal correction retained, Full CHAC achieves a similar SR of 29.60\% while reducing calls by 11.8\% and environment actions by 18.5\% relative to Fixed-5 + Terminal. SPL rises from 14.77\% to 15.93\%. The boundaries thus reduce inference and execution demand at a comparable observed success rate, while terminal correction accounts for most of the success recovery in this comparison. Individually removing Commitment or Visual from Full adds 5,300 or 1,052 calls. The following analyses examine this dependence on execution and test what information the history provides at stopping.

\subsection{Closed-Loop Boundary Coordination}
\label{sec:boundary-coordination}

\paragraph{Commitment-dependent cost reversal.}
The same visual refresh rule has opposite observed effects on total calls with Commitment off or on. With Terminal disabled, adding Visual increases calls by 1,297 under Fixed-5 but saves 1,052 with Commitment (Fig.~\ref{fig:boundary-results}(b)). Let $N_S$ denote total VLM calls with boundary set $S$. The call interaction is
\begin{equation}
\begin{aligned}
 \Delta_{C,V}&=(N_{C,V}-N_C)-(N_V-N_{\varnothing})\\
             &=-1{,}052-1{,}297=-2{,}349.
\end{aligned}
\label{eq:interaction}
\end{equation}
The 95\% interval from 20,000 paired bootstrap resamples is $[-4{,}108,-699]$ calls. Resampling 12 scene clusters and rescaling mean costs to 1,000 episodes gives $[-4{,}641,-366]$. Both intervals place the cost interaction below zero.

\begin{samepage}
Execution connects the two boundaries. A Commitment-triggered refresh renews both the prediction and the visual reference. The resulting actions change the observations evaluated by Visual. A Visual-triggered refresh changes the motion requests assessed by Commitment. The reversal is consistent with this dependence on intervening execution. Its evidence concerns cost: with Terminal off, adding Visual to Commitment changes SR from 23.10\% to 23.00\%.\par
\end{samepage}

\paragraph{Denser inference, lower total demand.}
Holding Visual and Terminal fixed, adding Commitment reduces calls from 35,111 to 29,811 (15.1\%) and actions from 119,236 to 95,328 (20.1\%). Logs record 3,652 Commitment refreshes and 8,952 fewer calls in other categories, a net reduction of 5,300. These trigger counts aggregate each configuration's realized trajectories. Meanwhile, inference density rises from 29.45 to 31.27 calls per 100 actions (6.2\%). With $N=Ar$, total calls combine action count $A$ and call density $r=N/A$. Here, the reduction in execution volume outweighs the increase in call density.

Figure~\ref{fig:boundary-results}(a) shows how this aggregate change distributes across episode pairs. Calls and actions both decrease in 31.9\% of pairs. Another 31.9\% require fewer actions without changing the number of calls. Overall, 74.6\% increase neither cost. Refresh decisions therefore affect both the density of inference and the amount of execution before termination.

Termination contributes materially to these savings. The 648 pairs in which both policies fail account for 3,619 of the 5,300 net saved calls (68.3\%). In 38 of these pairs, a timeout without Commitment becomes an earlier failed STOP with Commitment. Earlier termination of unsuccessful execution therefore contributes to the reduction in total demand.

\paragraph{Small displacement, changed visual evidence.}
In Fig.~\ref{fig:reorientation}, small displacement accompanies substantially changed visual evidence. In episode 1644, the robot is only 0.12\,m from the anchor but its heading differs by $129^\circ$, bringing a staircase into view. At their shared decision state, CHAC refreshes (Eq.~\eqref{eq:appearance}), while w/o Visual executes the cached fifth waypoint.

The subsequent action and observation sequences diverge. CHAC uses 41 VLM calls versus 47 and travels 20.64\,m less before terminal correction. The case connects directional evidence, refresh, and subsequent inference demand. Positions reached during execution remain available as history after the predictions are replaced.

\subsection{History-Guided Terminal Correction}
\label{sec:terminal-correction-results}

Table~\ref{tab:terminal} tests what visitation history contributes beyond the latest displacement. The three correction strategies share neural execution prefixes, requested correction radii, and native preview, budget, and STOP rules. Continuation follows the latest nonzero displacement, reversal flips it, and local consensus uses nearby visited positions. All use zero additional model calls. Evaluation includes all 1,000 episodes, including 41 timeouts.

\begin{table}[!ht]
\centering
\caption{\textbf{Terminal direction under matched requested radii.} All strategies share neural prefixes, native motor/STOP conditions, and zero additional VLM or Flow calls. SR includes all 1,000 episodes.}
\label{tab:terminal}
\small
\setlength{\tabcolsep}{4pt}
\renewcommand{\arraystretch}{1.15}
\begin{tabularx}{\columnwidth}{Xrrr}
\toprule
Strategy & Rescue $\uparrow$ & Regression $\downarrow$ & SR (\%) $\uparrow$\\
\midrule
Original STOP & 0 & 0 & 23.00\\
Motion continuation & 12 & 80 & 16.20\\
Motion reversal & 59 & 25 & 26.40\\
\rowcolor{gray!12}Local consensus & 82 & 16 & 29.60\\
\bottomrule
\end{tabularx}
\end{table}

Motion continuation reduces SR to 16.20\%, while reversal reaches 26.40\%. Local consensus achieves 29.60\%, with 82 rescues and 16 regressions, yielding 32 more successes than reversal. Under these matched conditions, local visitation geometry provides useful directional information beyond the latest displacement.

Among the 82 rescued episodes, 80 had previously recorded a goal distance below 0.25\,m before stopping outside that radius. The offline analysis uses goal distances to identify spatial progress lost before the final STOP. The correction uses visited positions. Nearby visited positions remain useful for terminal correction after the corresponding action predictions have expired. The correction adds at most one bounded movement, with zero additional VLM or Flow calls.

\subsection{Physical Execution}
We evaluate end-to-end navigation on a TurboPi robot with real camera observations, calibrated chassis control, and remote RTX 4090 inference. Table~\ref{tab:realworld} reports navigation outcomes and model inference costs under the same frozen policy.

CHAC achieves a success rate of 45.0\%, compared with 35.0\% for native execution and 40.0\% for Fixed-3. Relative to native execution, it reduces VLM calls from 687 to 310 and GPU inference time from 437.76\,s to 220.00\,s, corresponding to reductions of 54.9\% and 49.7\%, respectively. Compared with Fixed-3, CHAC improves success by 5.0 percentage points with 13.2\% fewer VLM calls and 16.2\% less GPU inference time. These aggregate results show improved navigation success with lower inference cost in the evaluated physical tasks.

\begin{table}[!ht]
\centering
\caption{\textbf{Physical navigation with a frozen policy.} Each method runs ten times on each of four indoor tasks. Success is reported as a percentage. Calls and GPU times are totals. GPU time measures model inference.}
\label{tab:realworld}
\small
\setlength{\tabcolsep}{3pt}
\begin{tabularx}{\columnwidth}{Xrrr}
\toprule
Method & Success (\%) & Calls & GPU (s)\\
\midrule
OmniNav-Fast & 35.0 & 687 & 437.76\\
Fixed-3 & 40.0 & 357 & 262.54\\
\rowcolor{gray!12}CHAC (Ours) & 45.0 & 310 & 220.00\\
\bottomrule
\end{tabularx}
\end{table}

\FloatBarrier
\section{Limitations}
CHAC reduces neural inference demand with task-dependent losses in success and SPL and increased environment actions. Physical latency depends on camera acquisition and executed motion. Terminal correction depends on the policy's STOP request and local visitation geometry. The physical trials use calibrated motion estimates and exercise prospective reuse. Paired simulation interventions provide the evidence for terminal recovery. Broader deployment with measured odometry would clarify motion robustness and end-to-end efficiency.

\section{Conclusion}
CHAC implements memory-guided adaptive reasoning with a frozen navigation policy. Retained visual context regulates prediction refresh, while visitation history enables bounded terminal correction without additional inference. Our analyses connect these decisions to navigation quality, inference demand, and physical execution costs.

{\small
\bibliographystyle{ieeenat_fullname}
\bibliography{references}
}
\clearpage
\appendix
\twocolumn[{
\begin{center}
{\Large\bfseries Appendix}
\end{center}
\section{Real-World Robot Evaluation}
\label{app:robot}

\begin{minipage}{\textwidth}
\centering
\includegraphics[width=\linewidth]{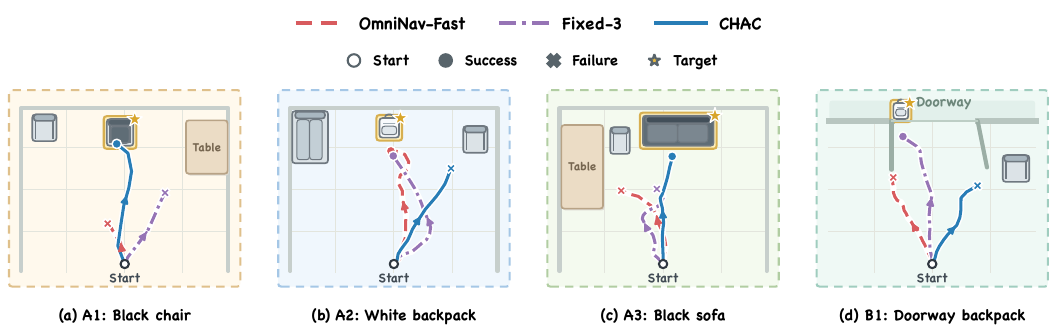}
\captionsetup{hypcap=false}
\captionof{figure}{\textbf{Navigation paths in four physical tasks.} Each panel shows the first run of OmniNav-Fast, Fixed-3, and CHAC on the corresponding task. Stars identify target objects; circles and crosses mark successful and unsuccessful endpoints, respectively. Table~\ref{tab:robot-detail} summarizes all ten runs per method and task.}
\label{fig:robot-tasks}
\end{minipage}
\vspace{12pt}
}]

\subsection{Platform and Deployment}
We use a TurboPi mobile robot with a Raspberry Pi 5, an onboard RGB camera, and a pan mechanism. The robot executes motion commands locally, while a remote RTX 4090 runs the frozen OmniNav-Fast policy. The evaluation configuration acquires left, front, and right images sequentially through a pan sweep. Each pan movement is configured for 400\,ms, followed by frame acquisition. The sweep is a single-camera hardware requirement. Three synchronized cameras covering the same views would eliminate camera rotation and settling delays without changing CHAC or the policy input interface. The three views and visual history condition the same waypoint-generation interface used by the navigation policy.

Native execution refreshes after its first waypoint. Fixed-3 executes up to three ordered waypoints per inference. CHAC regulates the cached prefix using its boundary checks. All three methods use the frozen model and the same task instructions. The physical adapter translates requested motion into calibrated chassis commands. Fixed-3 and CHAC use calibration-based motion estimates for their execution feedback.

\subsection{Tasks and Evaluation Protocol}
Figure~\ref{fig:robot-tasks} compares the paths from the first run of each method in the four task settings. A1--A3 specify direct approaches to a chair, backpack, and sofa. B1 specifies a backpack near a doorway.

\begin{table*}[t]
\centering
\caption{Robot navigation detail per task and method.}
\label{tab:robot-detail}
\small
\setlength{\tabcolsep}{5pt}
\renewcommand{\arraystretch}{1.12}
\begin{tabularx}{\textwidth}{ll*{5}{>{\raggedleft\arraybackslash}X}}
\toprule
Task & Method & Success & VLM Calls & GPU (s) & Actions & Reused\\
\midrule
A1 & OmniNav-Fast & 0/10 & 17.7 & 11.57 & 16.9 & 0.0\\
 & Fixed-3 & 3/10 & 6.4 & 4.90 & 16.5 & 11.0\\
\rowcolor{gray!12} & CHAC & 4/10 & 5.6 & 4.16 & 14.5 & 8.1\\
\midrule
A2 & OmniNav-Fast & 7/10 & 15.7 & 10.17 & 14.7 & 0.0\\
 & Fixed-3 & 7/10 & 8.8 & 6.08 & 23.4 & 15.6\\
\rowcolor{gray!12} & CHAC & 6/10 & 8.0 & 5.52 & 22.6 & 14.2\\
\midrule
A3 & OmniNav-Fast & 6/10 & 14.4 & 8.54 & 13.4 & 0.0\\
 & Fixed-3 & 0/10 & 10.3 & 7.29 & 29.6 & 19.5\\
\rowcolor{gray!12} & CHAC & 3/10 & 8.7 & 5.96 & 25.1 & 16.4\\
\midrule
B1 & OmniNav-Fast & 1/10 & 20.9 & 13.50 & 20.4 & 0.0\\
 & Fixed-3 & 6/10 & 10.2 & 7.98 & 27.9 & 18.5\\
\rowcolor{gray!12} & CHAC & 5/10 & 8.7 & 6.42 & 25.5 & 16.8\\
\bottomrule
\end{tabularx}
\end{table*}

The evaluation comprises 120 runs, with ten repetitions per method on each task. Runs use the task-specific marked starting locations and a 100-action budget. A1--A3 have a 180\,s time limit, and B1 has a 300\,s limit. Success requires reaching the intended target and issuing an autonomous STOP within the limit. Timeouts, manual takeovers, and unsuccessful stops count as failures. The tables summarize all repetitions.

Table~\ref{tab:realworld} reports aggregate inference costs. Table~\ref{tab:robot-detail} provides per-task success counts and mean inference and action statistics across all runs.

\subsection{Per-Task Results}
CHAC uses fewer VLM calls and less GPU inference time than both baselines on each of the four tasks. On A1, it succeeds in four of ten runs, compared with three for Fixed-3 and none for native execution, using 5.6 calls and 4.16\,s of GPU inference per run. On B1, it achieves five successes, compared with one for native execution and six for Fixed-3, while averaging 8.7 calls and 6.42\,s of GPU inference.

Success varies across tasks. On A2, CHAC succeeds in six runs, while both baselines succeed in seven. On A3, CHAC succeeds in three runs, compared with six for native execution and none for Fixed-3. Together, the four tasks yield 18 successes for CHAC, 14 for native execution, and 16 for Fixed-3, with lower aggregate inference cost for CHAC.

CHAC also executes fewer motion commands than Fixed-3 on all four tasks. This reduction accompanies its lower VLM-call counts and GPU inference costs, while the reported success counts reflect the same task-specific evaluation protocol.

\section{Mathematical Properties}
\label{app:math}
This section develops the equations in Section~\ref{sec:method} in the same order as the method. We first analyze the online horizon, the commitment and visual boundaries, and their joint effect on prediction reuse. We then derive the terminal projection, its spatial support and sensitivity, and its relationship to the native execution interface. All vector norms are Euclidean unless a subscript specifies otherwise.

\subsection{Online Action Reuse}
\label{app:online-properties}

\paragraph{Online horizon.}
Fix an inference index $k$ whose prediction does not request native STOP. We consider its boundary-defined horizon before applying the episode-budget cap. The information sets $\{\mathcal F_{k,j}\}_{j=0}^{H}$ form a filtration: $\mathcal F_{k,i}\subseteq\mathcal F_{k,j}$ for $i\le j$. At check $j$, the next action may use the observations, cached prediction, execution state, motor feedback, and read-only controller preview available at that check. Thus the index $j$ denotes a completed-waypoint count together with the information acquired before deciding whether to reuse the next waypoint. The first waypoint is handled by the native decoding and execution interface.

\paragraph{Proposition B.1 (Adapted first-event horizon).}
Suppose $C_{k,j},V_{k,j},M_{k,j}\in\mathcal F_{k,j}$ for $1\le j<H$, and let $B_{k,j}=C_{k,j}\cup V_{k,j}\cup M_{k,j}$. The horizon in Eq.~\eqref{eq:horizon} is a stopping time with values in $\{1,\ldots,H\}$.

\emph{Proof.}
For $1\le j<H$, a boundary-defined horizon ends by $j$ exactly when at least one boundary event occurs at a check no later than $j$. Hence
\begin{equation}
 \{h_k\le j\}=\bigcup_{i=1}^{j}B_{k,i}.
 \label{eq:app-horizon-event}
\end{equation}
Every $B_{k,i}$ in the union belongs to $\mathcal F_{k,i}\subseteq\mathcal F_{k,j}$, so the union is measurable with respect to $\mathcal F_{k,j}$. For $j=0$ the event is empty, and for $j\ge H$ it is the entire sample space. The convention $\inf\varnothing=+\infty$ gives $h_k=H$ when no boundary fires. This proves the assertion. Once reuse ends, its record can be held fixed for the remaining indices; no future observations are needed to identify the first event.\hfill$\square$

This statement formalizes the causal use of execution evidence. The episode budget can further truncate execution, and a native STOP request follows the terminal branch of Algorithm~\ref{alg:chac}.

\paragraph{Commitment boundary.}
In Eq.~\eqref{eq:commitment}, both $\mathbf u$ and $\widehat{\mathbf v}$ are horizontal displacements from the same current position. The requested waypoint lies at offset $\mathbf u$, while the controller preview reaches offset $\widehat{\mathbf v}$. We assume $\mathbf u\ne\mathbf0$, as required by the definition of $\gamma_{k,j}$.

\paragraph{Proposition B.2 (Geometry of supported progress).}
The controller preview satisfies
\begin{equation}
\begin{aligned}
 &\|\mathbf u-\widehat{\mathbf v}\|^2-\|\mathbf u\|^2\\
 &\qquad=\|\widehat{\mathbf v}\|^2
       -2\|\mathbf u\|\gamma_{k,j}.
\end{aligned}
\label{eq:app-progress-distance}
\end{equation}
Consequently, $\gamma_{k,j}\le0$ implies that the preview cannot reduce the distance to the requested waypoint. The distance increases strictly if $\widehat{\mathbf v}\ne\mathbf0$. More generally, the preview reduces this distance if and only if
\begin{equation}
 \gamma_{k,j}>
 \frac{\|\widehat{\mathbf v}\|^2}{2\|\mathbf u\|}.
 \label{eq:app-distance-descent}
\end{equation}

\emph{Proof.}
Expanding the squared distance gives
$\|\mathbf u-\widehat{\mathbf v}\|^2
=\|\mathbf u\|^2+\|\widehat{\mathbf v}\|^2
-2\mathbf u^\top\widehat{\mathbf v}$.
Substitution of Eq.~\eqref{eq:commitment} proves Eq.~\eqref{eq:app-progress-distance}. If $\gamma_{k,j}\le0$, both terms on its right-hand side are nonnegative, and the first is positive for nonzero $\widehat{\mathbf v}$. Solving the strict distance-decrease inequality yields Eq.~\eqref{eq:app-distance-descent}.\hfill$\square$

The zero threshold used by CHAC therefore detects a geometrically identifiable failure of the current motion request: the controller preview supplies no progress along that request and cannot approach its endpoint. Equation~\eqref{eq:app-distance-descent} also distinguishes positive directional progress from strict distance reduction; the implemented test remains $\gamma_{k,j}\le0$.

For a fixed request, the progress statistic has a direct perturbation bound. If two previews satisfy $\|\widehat{\mathbf v}'-\widehat{\mathbf v}\|\le\eta$, then
\begin{equation}
\begin{aligned}
 |\gamma'_{k,j}-\gamma_{k,j}|
 &=\frac{|\mathbf u^\top(\widehat{\mathbf v}'-\widehat{\mathbf v})|}{\|\mathbf u\|}
 \le\eta.
\end{aligned}
\label{eq:app-progress-stability}
\end{equation}
This follows from the Cauchy--Schwarz inequality. Thus $\gamma_{k,j}>\eta$ preserves positive progress under the perturbation, while $\gamma_{k,j}<-\eta$ preserves the nonpositive-progress trigger. The same distance identity applies to realized displacement $\mathbf v$ in the motor-feedback test.

The completion-score branch has a separate role. A cached score $c_k^j>0.5$ requests a new prediction before another cached MOVE. A single cached completion cue is distinct from the native STOP predicate, which requires all arrival scores of the evaluated prediction to exceed $0.5$. Accordingly, this branch changes when the prediction is refreshed while preserving the native stopping rule.

\paragraph{Visual boundary.}
We now analyze Eq.~\eqref{eq:appearance} at a fixed check $(k,j)$. The selected camera, the three-region partition, the action-region index $q^\star$, and the angular novelty predicate $\phi_t\in\Omega_t\setminus\Omega_k$ are held fixed. Region indices correspond between the two image collections being compared. The closest reference region is allowed to change.

\paragraph{Proposition B.3 (Stability of directional contrast).}
Let $R'_{t,q}$ and $A'_{k,s}$ be perturbed current and reference regions such that
\begin{equation}
\begin{aligned}
 \frac{\|R'_{t,q}-R_{t,q}\|_1}{P}&\le\varepsilon_R
 &&\text{for every }q,\\
 \frac{\|A'_{k,s}-A_{k,s}\|_1}{P}&\le\varepsilon_A
 &&\text{for every }s.
\end{aligned}
\label{eq:app-image-perturbation}
\end{equation}
Then the matched-region distances and directional contrast satisfy
\begin{equation}
\begin{aligned}
 |d'_{t,q}-d_{t,q}|&\le\varepsilon_R+\varepsilon_A,\\
 |\delta'_{k,j}-\delta_{k,j}|&\le2(\varepsilon_R+\varepsilon_A).
\end{aligned}
\label{eq:app-contrast-stability}
\end{equation}
Writing $\beta=2(\varepsilon_R+\varepsilon_A)$, the positive-contrast condition is preserved when $\delta_{k,j}>\beta$, and remains false when $\delta_{k,j}<-\beta$.

\emph{Proof.}
For each pair of regions, define the quantity already minimized in Eq.~\eqref{eq:appearance} as $D_{q,s}=\|R_{t,q}-A_{k,s}\|_1/P$. The reverse triangle inequality and Eq.~\eqref{eq:app-image-perturbation} give
\begin{equation}
\begin{aligned}
 |D'_{q,s}-D_{q,s}|
 &\le\frac{\|(R'_{t,q}-R_{t,q})-(A'_{k,s}-A_{k,s})\|_1}{P}\\
 &\le\varepsilon_R+\varepsilon_A.
\end{aligned}
\end{equation}
For two finite collections $\{a_s\},\{a'_s\}$ with $|a'_s-a_s|\le e$, each $a'_s$ lies between $a_s-e$ and $a_s+e$. Taking minima gives $|\min_s a'_s-\min_s a_s|\le e$. Taking maxima gives the analogous bound for maxima. Applying the minimum bound to $D_{q,s}$ establishes the first inequality in Eq.~\eqref{eq:app-contrast-stability}. Applying the maximum bound to the two non-action regions and then the triangle inequality yields
\begin{equation}
\begin{aligned}
 |\delta'_{k,j}-\delta_{k,j}|
 &\le |d'_{t,q^\star}-d_{t,q^\star}|\\
 &\quad+\left|\max_{q\ne q^\star}d'_{t,q}
                -\max_{q\ne q^\star}d_{t,q}\right|\\
 &\le2(\varepsilon_R+\varepsilon_A).
\end{aligned}
\end{equation}
The sign conclusions follow by adding or subtracting this bound.\hfill$\square$

The magnitude of the directional contrast thus supplies a margin for the appearance test. If both image collections have perturbation bound $\varepsilon$, the bound is $4\varepsilon$. Within a chunk, the cached reference is fixed; for perturbations to the current image alone it reduces to $2\varepsilon_R$. With the angular predicate fixed, these conclusions also determine whether the full visual event $V_{k,j}$ is preserved. A change in camera selection, action-region assignment, or angular novelty changes the inputs to this comparison and is treated as a different check.

\paragraph{Coordinated execution.}
The union $B_{k,j}=C_{k,j}\cup V_{k,j}\cup M_{k,j}$ connects the component tests to the horizon in Eq.~\eqref{eq:horizon}. At a fixed evidence record, ending reuse at the first satisfied predicate implements this union. Motor feedback obtained after a MOVE invalidates the cache before further reuse.

\paragraph{Proposition B.4 (Ordering under shared evidence).}
Consider two boundary families evaluated on the same indexed evidence sequence for the same prediction. If $B_{k,j}\subseteq B'_{k,j}$ for every $1\le j<H$, then
\begin{equation}
 h'_k\le h_k
 \quad\text{for every outcome}.
 \label{eq:app-horizon-order}
\end{equation}

\emph{Proof.}
If $h_k<H$, the original boundary fires at $j=h_k$. The enlarged boundary also fires at that index, if not earlier, so $h'_k\le h_k$. If $h_k=H$, the inequality follows from $h'_k\le H$.\hfill$\square$

The shared-evidence condition identifies the object being ordered: the lifetime of one prediction. A closed-loop refresh instead replaces that prediction and changes later actions and observations. For an episode with $A>0$ environment actions and $N$ model calls, let $d=N/A$ be the observed call density. For two runs with positive action and call counts,
\begin{equation}
 \frac{N'}{N}=\frac{d'}{d}\frac{A'}{A}.
 \label{eq:app-cost-decomposition}
\end{equation}
Hence $N'<N$ exactly when $A'/A<d/d'$. This accounting identity explains how higher call density can coexist with lower total inference demand when the executed episode is shorter. The boundary comparisons in Section~\ref{sec:boundary-coordination} evaluate these quantities on the realized runs and distinguish the recorded stopping outcomes.

\subsection{History-Guided Geometric Projection}
\label{app:projection-properties}

\paragraph{Records, coordinates, and objective.}
Fix a native STOP request at time $t$. Let $\mathbf y_1,\ldots,\mathbf y_n$ be the retained historical offsets, expressed in the same local horizontal frame as in Section~\ref{sec:terminal-projection}. The count $n$ includes repeated records. Add $\mathbf y_{n+1}=\mathbf y_{n+2}=\mathbf y_{n+3}=\mathbf0$ for the three current views. Thus $m=|\mathcal L_t|=n+3\ge3$, and
\begin{equation}
\begin{aligned}
 \overline{\mathbf y}_t&=\frac1{n+3}\sum_{i=1}^{n}\mathbf y_i,\\
 J_t(\boldsymbol\Delta)&=\frac1{n+3}
 \left(\sum_{i=1}^{n}\|\boldsymbol\Delta-\mathbf y_i\|^2
                         +3\|\boldsymbol\Delta\|^2\right).
\end{aligned}
\label{eq:app-terminal-objective}
\end{equation}
The second line is precisely the objective in Eq.~\eqref{eq:terminal}, with the three zero offsets written explicitly. The current views therefore contribute a quadratic penalty on displacement from the current position. Its coefficient follows from their native multiplicity. The candidate remains constrained to the disk $\mathbb B_\rho$ with the radius in Eq.~\eqref{eq:scale}.

\paragraph{Proposition B.5 (Projection and optimality gap).}
For $\rho\ge0$, Eq.~\eqref{eq:terminal} has a unique minimizer
\begin{equation}
 \boldsymbol\Delta_t^\star=
 \begin{cases}
 \overline{\mathbf y}_t,&\|\overline{\mathbf y}_t\|\le\rho,\\
 \displaystyle\rho\frac{\overline{\mathbf y}_t}{\|\overline{\mathbf y}_t\|},
   &\|\overline{\mathbf y}_t\|>\rho.
 \end{cases}
 \label{eq:app-projection-closed}
\end{equation}
It is characterized by
\begin{equation}
 \langle\overline{\mathbf y}_t-\boldsymbol\Delta_t^\star,
         \boldsymbol\Delta-\boldsymbol\Delta_t^\star\rangle\le0
 \quad(\boldsymbol\Delta\in\mathbb B_\rho).
 \label{eq:app-projection-vi}
\end{equation}
Every feasible candidate satisfies the quantitative optimality bound
\begin{equation}
 J_t(\boldsymbol\Delta)-J_t(\boldsymbol\Delta_t^\star)
 \ge\|\boldsymbol\Delta-\boldsymbol\Delta_t^\star\|^2.
 \label{eq:app-projection-gap}
\end{equation}

\emph{Proof.}
Using all $m=n+3$ offsets, expand each difference around the mean. Since $\sum_{i=1}^{m}(\mathbf y_i-\overline{\mathbf y}_t)=\mathbf0$, the cross terms cancel and
\begin{equation}
\begin{aligned}
 J_t(\boldsymbol\Delta)
 &=\|\boldsymbol\Delta-\overline{\mathbf y}_t\|^2+\sigma_t^2,\\
 \sigma_t^2&=\frac1m\sum_{i=1}^{m}
                    \|\mathbf y_i-\overline{\mathbf y}_t\|^2.
\end{aligned}
\label{eq:app-variance-decomposition}
\end{equation}
The dispersion $\sigma_t^2$ does not depend on the candidate. The objective is continuous and strictly convex, and the feasible disk is nonempty, compact, and convex, so its minimizer exists and is unique.

Write $r=\|\overline{\mathbf y}_t\|$. If $r\le\rho$, the mean itself is feasible and minimizes the squared-distance term. If $r>\rho$, the reverse triangle inequality gives $\|\boldsymbol\Delta-\overline{\mathbf y}_t\|\ge r-\rho$ for every feasible $\boldsymbol\Delta$, with equality attained at $\rho\overline{\mathbf y}_t/r$. This proves Eq.~\eqref{eq:app-projection-closed}, including the singleton disk when $\rho=0$.

For any feasible $\boldsymbol\Delta$, the segment $\boldsymbol\Delta_t^\star+s(\boldsymbol\Delta-\boldsymbol\Delta_t^\star)$ is feasible for $0\le s\le1$. The right derivative of the squared-distance objective at its minimum $s=0$ is nonnegative. Dividing this derivative by two gives
$\langle\boldsymbol\Delta_t^\star-\overline{\mathbf y}_t,
\boldsymbol\Delta-\boldsymbol\Delta_t^\star\rangle\ge0$,
which is Eq.~\eqref{eq:app-projection-vi}. Conversely, expansion gives
\begin{equation}
\begin{aligned}
 &J_t(\boldsymbol\Delta)-J_t(\boldsymbol\Delta_t^\star)\\
 &=\|\boldsymbol\Delta-\boldsymbol\Delta_t^\star\|^2\\
 &\quad+2\langle\boldsymbol\Delta_t^\star-\overline{\mathbf y}_t,
                 \boldsymbol\Delta-\boldsymbol\Delta_t^\star\rangle.
\end{aligned}
\label{eq:app-objective-gap-identity}
\end{equation}
The inner product is nonnegative by Eq.~\eqref{eq:app-projection-vi}, proving Eq.~\eqref{eq:app-projection-gap}. This also proves that the variational inequality characterizes the unique minimizer.\hfill$\square$

Taking $\boldsymbol\Delta=\mathbf0$ in Eq.~\eqref{eq:app-projection-gap} yields
\begin{equation}
 J_t(\mathbf0)-J_t(\boldsymbol\Delta_t^\star)
 \ge\|\boldsymbol\Delta_t^\star\|^2.
 \label{eq:app-zero-comparison}
\end{equation}
More explicitly, writing $r=\|\overline{\mathbf y}_t\|$, Eq.~\eqref{eq:app-projection-closed} gives
\begin{equation}
 J_t(\mathbf0)-J_t(\boldsymbol\Delta_t^\star)
 =\begin{cases}
 r^2,&r\le\rho,\\
 2\rho r-\rho^2,&r>\rho.
 \end{cases}
 \label{eq:app-exact-improvement}
\end{equation}
These expressions quantify the improvement in the historical squared-distance objective relative to leaving the current position unchanged.

\paragraph{Spatial support and current-position shrinkage.}
For $n>0$, let $\overline{\mathbf y}_{\mathrm{hist}}=n^{-1}\sum_{i=1}^{n}\mathbf y_i$. Equation~\eqref{eq:app-terminal-objective} gives
\begin{equation}
 \overline{\mathbf y}_t
 =\frac{n}{n+3}\overline{\mathbf y}_{\mathrm{hist}}.
 \label{eq:app-history-shrinkage}
\end{equation}
Thus the current-view records shrink the historical mean toward the current position. Historical duplicates retain their weight in this mean.

The mean $\overline{\mathbf y}_t$ is a convex combination of the historical offsets and zero. Equation~\eqref{eq:app-projection-closed} writes $\boldsymbol\Delta_t^\star=\alpha\overline{\mathbf y}_t$ with $\alpha\in[0,1]$; choose $\alpha=1$ if the mean is zero. Therefore
\begin{equation}
 \boldsymbol\Delta_t^\star
 \in\operatorname{conv}\{\mathbf0,\mathbf y_1,\ldots,\mathbf y_n\}
       \cap\mathbb B_\rho.
 \label{eq:app-spatial-support}
\end{equation}
Furthermore, locality filtering and horizontal projection give $\|\mathbf y_i\|\le Hb$ for each historical record. The triangle inequality then yields the tighter joint bound
\begin{equation}
 \|\boldsymbol\Delta_t^\star\|
 \le\min\left\{\rho,\frac{n}{n+3}Hb\right\}.
 \label{eq:app-joint-bound}
\end{equation}
For $n=0$, the mean and candidate are zero. The same is true for a nonempty history whose offsets sum to zero. Both cases preserve native STOP without a correction movement.

\paragraph{Proposition B.6 (Sensitivity to retained history).}
Consider two histories with the same retained-record indices and multiplicities, the same current pose and coordinate frame, and the same radius $\rho$. Let their historical offsets be $\mathbf y_i$ and $\mathbf y'_i$, and let their correction candidates be $\boldsymbol\Delta_t^\star$ and $\boldsymbol\Delta_t^{\star\prime}$. Then
\begin{equation}
\begin{aligned}
 \|\boldsymbol\Delta_t^\star-\boldsymbol\Delta_t^{\star\prime}\|^2
 &\le\langle\boldsymbol\Delta_t^\star-\boldsymbol\Delta_t^{\star\prime},
       \overline{\mathbf y}_t-\overline{\mathbf y}'_t\rangle,\\
 \|\boldsymbol\Delta_t^\star-\boldsymbol\Delta_t^{\star\prime}\|
 &\le\|\overline{\mathbf y}_t-\overline{\mathbf y}'_t\|\\
 &\le\frac1{n+3}\sum_{i=1}^{n}\|\mathbf y_i-\mathbf y'_i\|.
\end{aligned}
\label{eq:app-history-stability}
\end{equation}
In particular, if each historical offset changes by at most $\varepsilon$, the candidate changes by at most $n\varepsilon/(n+3)$.

\emph{Proof.}
Use $\boldsymbol\Delta_t^{\star\prime}$ as the feasible comparison point in Eq.~\eqref{eq:app-projection-vi}, and then exchange the two histories and candidates. The resulting inequalities are
\begin{equation}
\begin{aligned}
 \langle\overline{\mathbf y}_t-\boldsymbol\Delta_t^\star,
         \boldsymbol\Delta_t^{\star\prime}-\boldsymbol\Delta_t^\star\rangle&\le0,\\
 \langle\overline{\mathbf y}'_t-\boldsymbol\Delta_t^{\star\prime},
         \boldsymbol\Delta_t^\star-\boldsymbol\Delta_t^{\star\prime}\rangle&\le0.
\end{aligned}
\end{equation}
Adding and rearranging proves the first inequality in Eq.~\eqref{eq:app-history-stability}. If the candidates differ, apply Cauchy--Schwarz and divide by their distance to obtain the second; if they coincide it holds immediately. The last bound follows by subtracting the two means in Eq.~\eqref{eq:app-terminal-objective} and applying the triangle inequality.\hfill$\square$

The fixed-membership condition can itself be checked through a locality margin. Consider a fixed pool of candidate historical records, with current pose and $Hb$ unchanged. If every candidate position changes by at most $\varepsilon$ and every original distance satisfies
\begin{equation}
 \big|\|\mathbf p_i-\mathbf p_t\|-Hb\big|>\varepsilon,
 \label{eq:app-locality-margin}
\end{equation}
then no record crosses the locality cutoff. Indeed, the reverse triangle inequality bounds the change in each distance by $\varepsilon$. Rotation into the fixed local frame preserves norms, and horizontal projection cannot increase them, so the corresponding offset changes are also at most $\varepsilon$. Proposition B.6 therefore applies to the complete filtering-and-projection calculation under this margin. The condition is imposed on both retained and excluded records in the fixed pool.

\paragraph{Connection to native execution.}
The projection defines a requested displacement. Algorithm~\ref{alg:chac} converts it to the native distance-and-heading command, checks positive previewed progress, and executes at most one correction MOVE before STOP. If the candidate is zero, the preview does not support positive progress, or fewer than two action slots remain, the terminal branch executes STOP directly. The decision uses the existing history, arithmetic projection, and controller preview, adding zero neural evaluations. An accepted correction uses one additional movement slot relative to direct STOP; the two-slot requirement accounts for that MOVE and the original STOP.

The relation between a requested candidate and a realized displacement can be expressed using the same objective. Hold the terminal history and its coordinate frame fixed. Suppose the realized horizontal displacement is $\mathbf v=\boldsymbol\Delta_t^\star+\mathbf e$, where $\|\mathbf e\|\le\eta$. Equation~\eqref{eq:app-variance-decomposition} gives the exact identity
\begin{equation}
\begin{aligned}
 J_t(\mathbf v)-J_t(\boldsymbol\Delta_t^\star)
 =\|\mathbf e\|^2
 +2\langle\boldsymbol\Delta_t^\star-\overline{\mathbf y}_t,\mathbf e\rangle.
\end{aligned}
\label{eq:app-realization-identity}
\end{equation}
Since $\|\boldsymbol\Delta_t^\star-\overline{\mathbf y}_t\|=\max(r-\rho,0)$, this yields
\begin{equation}
 J_t(\mathbf v)-J_t(\boldsymbol\Delta_t^\star)
 \le\eta^2+2\max(r-\rho,0)\eta.
 \label{eq:app-realization-bound}
\end{equation}
The bound follows from Cauchy--Schwarz and applies even if the realized displacement lies outside the requested disk. Together with Eq.~\eqref{eq:app-exact-improvement}, it gives a sufficient condition for realized improvement in the historical objective: its right-hand side must be smaller than $J_t(\mathbf0)-J_t(\boldsymbol\Delta_t^\star)$. This explicitly connects projection quality to motion-execution accuracy. The positive-progress check assesses the controller preview; an error bound $\eta$ describes the separate relation between the request and the realized motion. Navigation outcomes of terminal correction are evaluated by the matched interventions in Section~\ref{sec:terminal-correction-results}.

\begin{figure*}[!t]
\centering
\includegraphics[width=\textwidth]{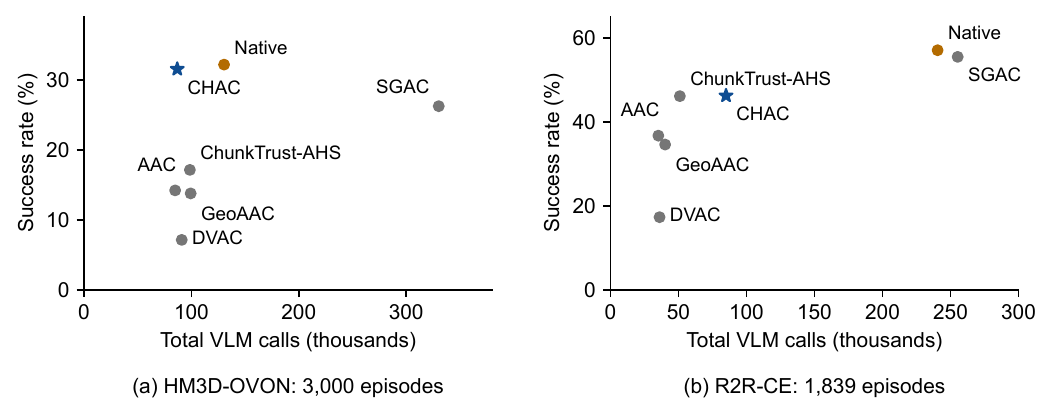}
\caption{\textbf{Success rate and VLM calls.} All methods and values come from Table~\ref{tab:main}. Blue stars mark CHAC; orange points mark native execution. Moving toward the upper left means higher success with fewer VLM calls. Each point represents one evaluated method configuration.}
\label{fig:tradeoff}
\end{figure*}

\section{Implementation Details}
\label{app:implementation}
\paragraph{Cache and boundary checks.}
CHAC stores the five-waypoint prediction, its inference-frame reference, and a cursor into the ordered chunk. Continuation decodes the next cached waypoint in the current execution frame. Visual, task, and execution checks run in this order and stop at the first trigger. A refresh discards the remaining cached waypoints and establishes a new visual reference. Motor feedback can invalidate the cache after execution. The model weights, waypoint order, and native decoding rules remain fixed.

\paragraph{Visual evidence.}
The simulation implementation caches three grayscale references at $32\times32$ resolution. Comparisons use $32\times30$ images split into three vertical regions. Each of the three current regions is compared with nine reference regions, yielding a fixed-size pixel calculation per check. Angular coverage includes the inference views and native visual history. Command odometry updates the current action direction. The image reference stays unchanged until the next policy invocation.

\paragraph{Motion interfaces.}
Simulation previews use the native controller's filtered and snapped displacement while preserving the simulator state. Physical deployment acquires the three views sequentially and executes calibrated chassis commands. Its feedback and projection interface use motion estimates from the calibration. Both adapters supply motion feedback to the execution checks.

\paragraph{Terminal records.}
The simulation terminal stage reads position records aligned with native visual history and the three current views. It retains multiplicity, applies the $Hb$ locality radius, and clips the mean offset to $\rho=\min(b,s_{\mathrm{motor}})$. For the reported simulation setting, $H=5$, $b\approx0.424$\,m, and $s_{\mathrm{motor}}=0.25$\,m. The local history radius is approximately 2.12\,m. A single pass over retained records computes the mean, followed by a constant-size projection and native admissibility check.

\paragraph{HM3D-OVON baseline scoring.}
For AAC in Table~\ref{tab:main}, we rescore the recorded trajectories using the common success criterion of an actual STOP at a goal distance below $0.25\,\mathrm{m}$.

\section{Computational Resources}
Simulation experiments use four NVIDIA RTX PRO 5000 GPUs with 72\,GB of memory per GPU. Physical-robot inference uses one NVIDIA RTX 4090, with image capture and chassis execution on the Raspberry Pi 5. The policy remains frozen throughout both settings. Each standard invocation uses one VLM forward and ten flow-head updates. Main benchmark tables account for VLM calls and flow-update equivalents, including additional sampling by the compared methods. The physical deployment additionally reports GPU inference time and executed and reused actions as per-run means in Table~\ref{tab:robot-detail}.
\section{Quality and Inference Operating Points}
\label{app:tradeoff}

The two tasks expose different operating trade-offs. On HM3D, CHAC retains near-native success at fewer VLM calls. On R2R, ChunkTrust-AHS achieves similar success at fewer calls than CHAC. Table~\ref{tab:main} complements these operating points with SPL, flow-update equivalents, and environment actions.

\end{document}